\documentclass[runningheads]{llncs}

\usepackage{eccv}

\usepackage{eccvabbrv}

\usepackage{graphicx}
\usepackage{booktabs}

\usepackage[accsupp]{axessibility}  
\usepackage{comment}

\usepackage{orcidlink}

\begin{document}

\title{Unified Driving Tokens: Representation- and Geometry-Guided Discrete Tokenizer for Driving World Models and Planning}

\titlerunning{Unified Driving Tokens}
\author{Ziyang Yao\inst{1,2} \and Zeyu Zhu\inst{2}\and YunCheng Jiang\inst{2}\and Zibin Guo\inst{2}\and Huijing Zhao\inst{1}}

\authorrunning{Z.~Yao et al.}

\institute{Peking University \and
Xiaomi EV}

\maketitle

\begin{abstract}
Discrete visual tokens should provide a compact representation for both token-based world modeling and planning in autonomous driving. However, most tokenizers are inherited from image generation and are optimized mainly for pixel reconstruction, which may leave a gap between what is easy to generate and what is useful to decode for driving decisions. We present a representation-guided and geometry-enhanced tokenizer that learns discrete tokens under joint supervision. The tokenizer aligns its discrete bottleneck with a frozen DINO feature space through feature decoding, while preserving appearance via RGB reconstruction with perceptual and adversarial losses. To inject geometric state-related cues, we add adjacent-frame depth and relative-pose supervision during training and stabilize joint objectives with multi-codebook quantization. We evaluate the same learned tokens with a lightweight planning readout and a GPT-style next-token world model. Experiments on NAVSIM show improved reconstruction fidelity and representation consistency, competitive planning performance under a fixed decoder, and better generative quality under matched settings.

\end{abstract}

\section{Introduction}
\label{sec:intro}

Planning and decision-making are central problems in autonomous driving: given sensory observations and past actions, the system must produce safe, comfortable, and efficient future trajectories~\cite{dataset2024.1,dataset2021.1,e2e2023.1,sur2024.1}.
Recent progress increasingly leverages generative world models to support planning, by learning the conditional distribution of environment evolution for future prediction, counterfactual imagination, and controllable data synthesis~\cite{e2e2025.2,e2e2025.4,e2e2025.5,e2e2025.6,sur2025.1}.
A particularly scalable instantiation casts driving as sequence modeling: mapping observations (optionally conditioned on actions and language) into discrete token sequences and training with next-token prediction~\cite{baseline2025.2,e2e2025.4}.
In this paradigm, a discrete visual tokenizer is not merely a compressor; it defines the discrete language of the driving world—a shared interface that must be amenable to generative sequence learning while being directly consumable by planning and decision modules.

Most existing discrete tokenizers are inherited from image generation, where they are optimized to support two-stage token-based synthesis and to preserve perceptual fidelity under pixel reconstruction~\cite{dl2017.1,tok2024.1}.
For autonomous driving, however, discrete tokens must play a dual role: they are the prediction targets for token-based world modeling and the input representation that downstream planners decode~\cite{baseline2025.2}.
Existing driving token pipelines often optimize tokenizers primarily for pixel reconstruction and generation, and assess planning performance only after the fact, rather than treating planning-consumability as a first-class objective during token learning~\cite{e2e2025.5,e2e2025.4,baseline2025.2}.
Motivated by making discrete tokens more useful beyond pure reconstruction, prior work in image/video generation has explored enriching tokenizer supervision beyond pixels—e.g., aligning discrete tokens to strong pre-trained representations via feature reconstruction or distillation—so that tokens carry higher-level structure that downstream modules can decode and exploit, e.g., vision-language tasks~\cite{tok2025.1,tok2025.2}.
For driving, the requirement is sharper: tokens should preserve appearance for generation, encode semantics for scene understanding, and capture geometry/motion cues for planning—suggesting a need for a unified discrete interface that supports both planning consumption and world model generation.
Yet, richer alignment also stresses VQ quantization—leading to information loss and codebook instability (collapse and low utilization)—and may degrade the appearance fidelity that generative models rely on~\cite{tok2025.1}.

We therefore propose a representation-guided and geometry-enhanced discrete tokenizer for autonomous driving, explicitly designed as a shared token interface for both token-based world modeling and planning consumption.
Our tokenizer builds on a VQ-VAE~\cite{dl2017.1} backbone with codebook quantization to produce discrete tokens.
On the encoder side, we use frozen DINO \cite{dl2025.1} features as the primary input to provide instance semantic information that autonomous driving is concerned with, and introduce an explicit image-detail branch to recover textures and boundaries weakened by high-level features and quantization~\cite{tok2025.3}.
On the decoder side, we jointly reconstruct RGB and DINO features, constraining discrete tokens to approximate a strong semantic representation while retaining generative appearance.
To inject state-aware cues for planning, we further decode depth and ego-motion from quantized representations of adjacent frames, enabling temporal geometric supervision that complements semantic alignment~\cite{dl2025.2}.
Finally, we adopt multi-codebook quantization to alleviate capacity bottlenecks under joint supervision and to stabilize training~\cite{tok2025.1, tok2022.1}.
To evaluate planning consumability, we train a lightweight planning decoder on frozen discrete tokens to predict future trajectories.
To evaluate generative ability, we train an autoregressive next-token Transformer world model on token sequences  and assess generation quality.

Our contributions are fourfold.

(\textbf{1}) We propose a unified discrete-token learning framework for autonomous driving, viewing tokens as the {discrete driving language} that simultaneously serves as the modeling target for token-based world models and the consumable input representation for planning.

(\textbf{2}) We design a semantic representation-guided discrete VQ tokenizer that aligns tokens with strong pre-trained representations while preserving appearance fidelity, via frozen-representation encoding, an explicit detail pathway, and joint reconstruction of RGB and semantic representation features.

(\textbf{3}) We further make the discrete tokens geometry-aware and quantization-stable by introducing temporal depth and ego-motion supervision and multi-codebook quantization, enabling a controlled balance among appearance, semantics, and geometry under joint training.

(\textbf{4}) We demonstrate planning benefits of the learned discrete interface by plugging the same tokens into a lightweight planner and an autoregressive next-token Transformer world model, improving planning metrics  and generation quality against representative baselines under matched settings.

\section{Related Work}
\subsection{Discrete Tokenizer for Visual Generation}
Recent progress in visual generation has been largely driven by continuous paradigms, especially diffusion models, which offer strong fidelity when operating in pixel space or learned continuous latents \cite{ho2020ddpm,rombach2022ldm,peebles2023dit}. In parallel, discrete generation revisits image synthesis as sequence modeling by compressing images into a finite token vocabulary, typically via codebook-based tokenizers such as VQ-VAE and VQGAN \cite{vanDenOord2017vqvae,esser2021vqgan}. This tokenization interface makes visual synthesis compatible with autoregressive and masked token modeling, enabling scalable training and efficient sampling \cite{chang2022maskgit,tok2022.1}. More importantly, recent results indicate that, with sufficiently strong tokenizers and recipes, token-based autoregressive generation can be competitive with diffusion and even outperform it in certain scaling regimes \cite{tok2024.1,tok2025.3}. These trends suggest that discrete tokenizers are not merely an engineering convenience, but a principled representation layer that unifies reconstruction and generation while providing a natural bridge to large sequence models, which is increasingly important for scalable visual generation \cite{tok2025.1,tok2025.2}.
\subsection{World Model–Driven Planning}

A recent trend in autonomous driving is to couple world modeling and planning at the model level, rather than using world models only as data generators or simulators. Representative works unify state-action-conditioned scene evolution with downstream decision making, including autoregressive diffusion world models \cite{e2e2025.2}, policy-centric world modeling for collaborative state-action prediction \cite{e2e2025.4}, scaling analyses showing world models can amplify data scaling effects \cite{e2e2025.5}, and formulations that explicitly bridge planning with video generation in a latent driving world \cite{e2e2025.6}. In the same spirit, earlier efforts cast driving world modeling as controllable sequence generation and enable planning over multiple futures \cite{hu2023gaia1,wang2024drivewm}, and more recent end-to-end frameworks incorporate intention-aware latent world representations for trajectory evaluation and selection \cite{zheng2025world4drive}. While these methods often borrow core techniques from image or vision-language generation and then tailor objectives and conditioning for driving-specific constraints, the visual tokenizer, especially the discrete tokenizer, is still commonly pretrained with reconstruction-oriented objectives inherited from generic visual generation \cite{vanDenOord2017vqvae,esser2021vqgan}. This leaves a gap between token semantics optimized for generic reconstruction and the requirements of safety-critical planning, suggesting that learning tokenizers explicitly suited for autonomous driving is a key open problem.

\section{Method}
\label{sec:method}

\begin{figure}[tb]
  \centering
  \includegraphics[width=\linewidth, trim=0mm 50mm 0mm 0mm, clip]{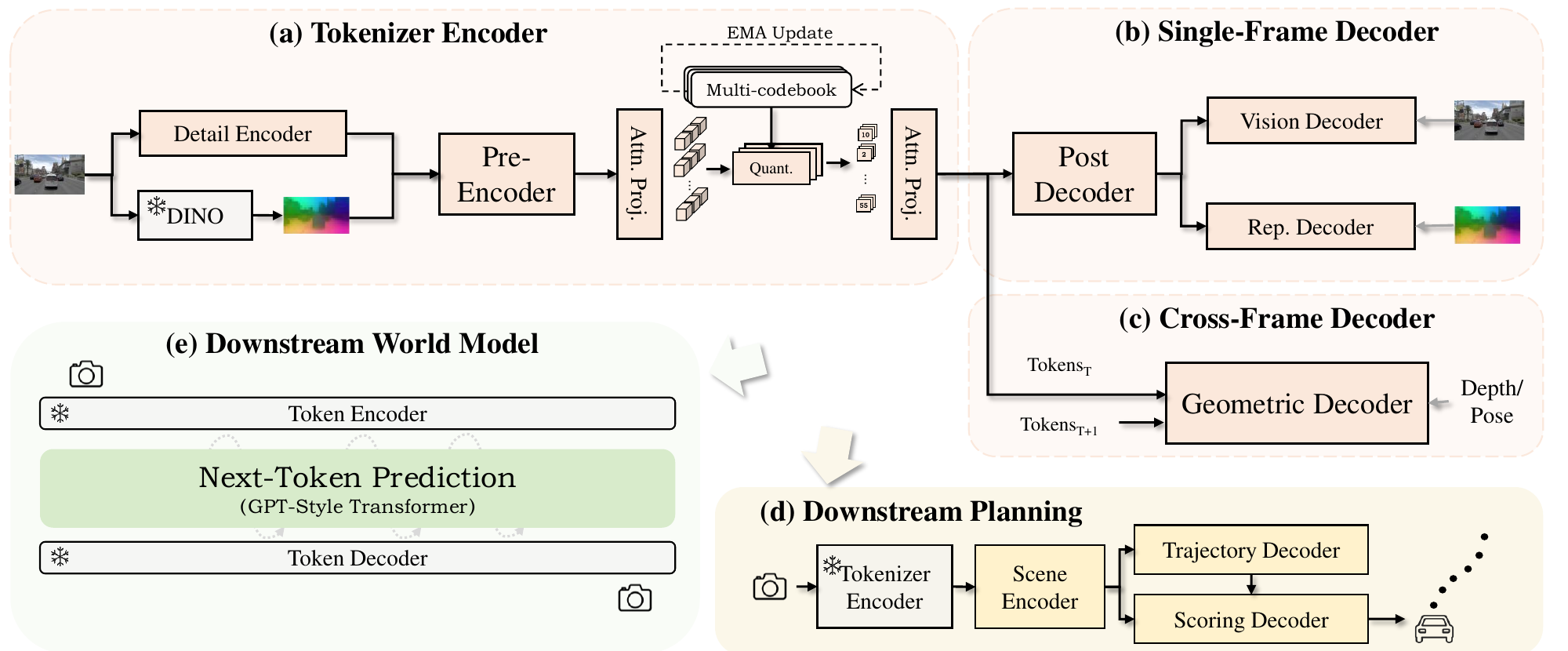}
  \caption{
   The overall framework of our method. Subfigures (a), (b), and (c) illustrate the overall architecture of the discrete tokenizer, while subgraphs (d) and (e) introduce the two downstream consumption tasks.
  }
  \label{fig:2}
\end{figure}

\subsection{Overview}
\label{sec:method_overview}

We study discrete visual tokenization for autonomous driving, where the learned tokens must support two downstream uses discussed in \cref{sec:intro}: they should be easy to model for token-based world model generation, and they should serve as effective inputs for planning.

Given an RGB frame $\mathbf{I}_t\in\mathbb{R}^{H\times W\times 3}$, the tokenizer outputs a grid of discrete token indices $\mathbf{k}_t\in\{1,\dots,K\}^{L}$ with corresponding embeddings $\mathbf{E}_t\in\mathbb{R}^{L\times d_q}$.
These tokens are trained to preserve appearance fidelity, capture semantic representations aligned with a frozen foundation model, and encode geometric cues through cross-frame supervision.
They can be decoded to RGB and semantic representations for reconstruction objectives, and they provide a compact discrete bottleneck that can be consumed by downstream models.

We first describe the representation-guided and geometry-enhanced tokenizer that learns the discrete tokens.
We then assess whether the tokens are usable for planning by training a lightweight trajectory decoder on frozen tokens.
Finally, we train an autoregressive next-token Transformer world model on the same token sequences to evaluate generative modelability and rollout quality.

\subsection{Representation-Guided and Geometry-Enhanced Tokenizer}
\label{sec:tokenizer}

\subsubsection{Preliminaries and Notation}
\label{sec:tok_notation}
We partition he RGB frame from the front camera $\mathbf{I}_t$ into non-overlapping patches of size $P\times P$, yielding a patch grid of $H_p=\frac{H}{P}$ and $W_p=\frac{W}{P}$ with $L=H_pW_p$ patch locations.
We use $t$ to index time and $l\in\{1,\dots,L\}$ to index patch locations.

We denote a frozen foundation visual model as $\Phi(\cdot)$ (e.g., DINO~\cite{dl2025.1}), which outputs normalized patch-level features:
\begin{equation}
\mathbf{F}_t=\Phi(\mathbf{I}_t)\in\mathbb{R}^{L\times d_{\text{dino}}}.
\label{eq:dino_feats}
\end{equation}
We treat $\Phi$ as fully frozen and do not back-propagate gradients through it.

\subsubsection{Tokenizer Architecture}
\label{sec:tok_arch}

We build a discrete tokenizer that preserves appearance fidelity while aligning tokens to a semantic representation extracted by a frozen foundation model.
The encoder combines an RGB detail branch with patch-wise DINO features, and encodes the fused sequence with a pre-norm Transformer \cite{dl2017.2} using Rotary Positional Embeddings (RoPE)~\cite{dl2021.1} to model spatial ordering on the patch grid.

We first extract patch embeddings from the RGB frame with a lightweight encoder $E_{\text{rgb}}$:
\begin{equation}
\mathbf{R}_t = E_{\text{rgb}}(\mathbf{I}_t)\in\mathbb{R}^{L\times d_{\text{rgb}}}.
\label{eq:rgb_tokens}
\end{equation}
We then concatenate $\mathbf{R}_t$ with frozen DINO patch features $\mathbf{F}_t$ and project them into a shared hidden space before applying a global Transformer encoder $E_{\text{g}}$:
\begin{equation}
\mathbf{X}_t = W_f[\mathbf{R}_t;\mathbf{F}_t]\in\mathbb{R}^{L\times d},
\qquad
\mathbf{H}_t = E_{\text{g}}(\mathbf{X}_t)\in\mathbb{R}^{L\times d}.
\label{eq:fuse_global}
\end{equation}

The discrete bottleneck is obtained with vector quantization.
We project $\mathbf{H}_t$ to the quantization space and assign each patch token to its nearest codeword in a codebook $\mathcal{C}=\{\mathbf{c}_k\}_{k=1}^{K}$:
\begin{align}
\mathbf{Z}_t &= P_{\text{in}}(\mathbf{H}_t)\in\mathbb{R}^{L\times d_q}, \label{eq:pre_vq_short}\\
k_{t,l} &= \arg\min_{k\in\{1,\dots,K\}}\|\mathbf{z}_{t,l}-\mathbf{c}_k\|_2^2,
\qquad
\mathbf{e}_{t,l}=\mathbf{c}_{k_{t,l}}. \label{eq:vq_assign}
\end{align}
We denote the discrete index map as $\mathbf{k}_t=\{k_{t,l}\}_{l=1}^{L}$ and the quantized embedding sequence as $\mathbf{E}_t=\{\mathbf{e}_{t,l}\}_{l=1}^{L}$.

For decoding, we map $\mathbf{E}_t$ back to the model dimension and apply a shared post-Transformer $E_{\text{post}}$ (also with RoPE).
Two lightweight decoders then reconstruct the RGB image and the semantic representation in the DINO feature space:
\begin{align}
\tilde{\mathbf{H}}_t &= P_{\text{out}}(\mathbf{E}_t),
\qquad
\mathbf{S}_t = E_{\text{post}}(\tilde{\mathbf{H}}_t), \label{eq:post_transformer}\\
\hat{\mathbf{I}}_t &= D_{\text{img}}(\mathbf{S}_t),
\qquad
\hat{\mathbf{F}}_t = D_{\text{dino}}(\mathbf{S}_t). \label{eq:decode_heads}
\end{align}
Reconstructing $\hat{\mathbf{F}}_t$ encourages the discrete tokens to retain semantic representations that downstream models can decode, while $\hat{\mathbf{I}}_t$ preserves the appearance information required by generative modeling.

\subsubsection{Training Objectives}
\label{sec:tok_loss}

We train the tokenizer to balance appearance fidelity, semantic representation alignment, and a stable discrete bottleneck.
The overall objective is a weighted sum of reconstruction, semantic alignment, adversarial training, and quantization regularization:
\begin{equation}
\mathcal{L}_{\text{tok}}
=
\lambda_{\text{rec}}\,\mathcal{L}_{\text{rec}}
+
\lambda_{\text{sem}}\,\mathcal{L}_{\text{sem}}
+
\lambda_{\text{gan}}\,\mathcal{L}_{\text{gan}}
+
\lambda_{\text{vq}}\,\mathcal{L}_{\text{vq}} .
\label{eq:tok_loss_simple}
\end{equation}

The appearance reconstruction loss $\mathcal{L}_{\text{rec}}$ combines a pixel-wise $\ell_2$ term with an LPIPS perceptual term:
\begin{equation}
\mathcal{L}_{\text{rec}}
=
\|\hat{\mathbf{I}}_t-\mathbf{I}_t\|_2^2
+
\lambda_{\text{lpips}}\,\mathrm{LPIPS}(\hat{\mathbf{I}}_t,\mathbf{I}_t).
\label{eq:rec_loss}
\end{equation}
Semantic representation alignment is imposed by reconstructing DINO patch features, using the sum of a cosine-similarity term and an MSE term between $\hat{\mathbf{F}}_t$ and $\mathbf{F}_t$; we denote their combination as $\mathcal{L}_{\text{sem}}$ for brevity.
We introduce a discriminator $D$ and optimize the generator with a standard GAN loss $\mathcal{L}_{\text{gan}}$ to distinguish $\mathbf{I}_t$ from $\hat{\mathbf{I}}_t$.

\subsubsection{Vector Quantization}
\label{sec:vq}

To support stable multi-objective training under appearance reconstruction, semantic representation alignment, and geometric supervision, We use a hard vector-quantization bottleneck with an exponential-moving-average (EMA)-updated codebook.
Let $\mathbf{Z}\in\mathbb{R}^{B\times L\times d_q}$ denote pre-quantization features (with $d_q=64$) and let $\mathbf{E}=\{\mathbf{e}_k\}_{k=1}^{K}$ be a single codebook with $K=16384$.

Each token is deterministically assigned to its nearest codeword by Euclidean distance:
\begin{equation}
I_{b,l}=\arg\min_{k\in\{1,\dots,K\}}\|\mathbf{z}_{b,l}-\mathbf{e}_k\|_2^2,
\qquad
\mathbf{q}_{b,l}=\mathbf{e}_{I_{b,l}}.
\label{eq:vq_hard_main}
\end{equation}
We use a straight-through estimator so that the forward pass uses $\mathbf{Q}$ while gradients can flow to the encoder:
\begin{equation}
\tilde{\mathbf{Q}}=\mathbf{Z}+\mathrm{sg}(\mathbf{Q}-\mathbf{Z}),
\label{eq:vq_ste_main}
\end{equation}
where $\mathbf{Q}$ stacks $\mathbf{q}_{b,l}$ and $\mathrm{sg}(\cdot)$ denotes stop-gradient.

We further include a commitment term that pulls $\mathbf{Z}$ toward the selected codewords:
\begin{equation}
\mathcal{L}_{\text{commit}}
=
\lambda_{c}\cdot
\frac{1}{BL}\sum_{b,l}
\left\|
\mathbf{z}_{b,l}-\mathrm{sg}(\mathbf{q}_{b,l})
\right\|_2^2.
\label{eq:vq_commit_main}
\end{equation}

Compared to the standard VQ-VAE formulation that learns codewords by gradient updates, our codebook is updated primarily by an exponential moving average rule that tracks cluster statistics induced by the hard assignments.
This choice decouples codeword updates from the competing multiple objectives and empirically improves stability under joint supervision.
In addition, we apply dead-code reinitialization and a weak orthogonality regularizer over active codewords to improve utilization and reduce redundancy.

\subsubsection{Cross-Frame Geometric Supervision}
\label{sec:geo_sup}

Frame-wise reconstruction provides only per-image constraints and does not explicitly encourage tokens to carry temporally consistent geometric cues.
We therefore introduce a geometry branch used only during training, inspired by VGGT-style spatiotemporal token aggregation~\cite{dl2025.2}.
In our setting, the cross-frame supervision is applied on adjacent frames only.

Given two adjacent frames $\mathbf{I}_{t}$ and $\mathbf{I}_{t+1}$, we obtain post-quantization token features from the tokenizer,
denoted by $\tilde{\mathbf{H}}_{t},\tilde{\mathbf{H}}_{t+1}\in\mathbb{R}^{L\times d}$.
We append an ego token $\mathbf{g}_{t},\mathbf{g}_{t+1}\in\mathbb{R}^{d}$ and apply a temporal aggregator $A_{\psi}$ that alternates
frame-wise attention and cross-frame attention to produce geometry-aware patch tokens $\mathbf{U}_{t},\mathbf{U}_{t+1}\in\mathbb{R}^{L\times d}$
and updated ego tokens $\bar{\mathbf{g}}_{t},\bar{\mathbf{g}}_{t+1}\in\mathbb{R}^{d}$:
\begin{equation}
\{\mathbf{U}_{t},\bar{\mathbf{g}}_{t},\mathbf{U}_{t+1},\bar{\mathbf{g}}_{t+1}\}
=
A_{\psi}\!\left(\{[\mathbf{g}_{t};\tilde{\mathbf{H}}_{t}],[\mathbf{g}_{t+1};\tilde{\mathbf{H}}_{t+1}]\}\right).
\label{eq:geo_agg_pair}
\end{equation}

Depth is decoded from $\mathbf{U}_{t}$ and $\mathbf{U}_{t+1}$ using a DPT-style dense prediction head that also outputs a confidence map:
\begin{equation}
\hat{\mathbf{D}}_{t},\,\hat{\mathbf{C}}_{t} = D_{\omega}(\mathbf{U}_{t}),
\qquad
\hat{\mathbf{D}}_{t+1},\,\hat{\mathbf{C}}_{t+1} = D_{\omega}(\mathbf{U}_{t+1}).
\label{eq:depth_head_pair}
\end{equation}
We regress the relative pose $\hat{\mathbf{p}}_{t\rightarrow t+1}=[\hat{\mathbf{t}}_{t\rightarrow t+1},\hat{\mathbf{q}}_{t\rightarrow t+1}]$
from the updated ego tokens:
\begin{equation}
\hat{\mathbf{p}}_{t\rightarrow t+1} = h_{\eta}\!\left([\bar{\mathbf{g}}_{t};\bar{\mathbf{g}}_{t+1}]\right).
\label{eq:pose_head_pair}
\end{equation}

We supervise the relative pose with an $\ell_1$ translation loss and a sign-invariant quaternion loss:
\begin{equation}
\mathcal{L}_{\text{pose}}
=
\left\|\hat{\mathbf{t}}_{t\rightarrow t+1}-\mathbf{t}_{t\rightarrow t+1}\right\|_1
+
\min\!\left(\left\|\hat{\mathbf{q}}_{t\rightarrow t+1}-\mathbf{q}_{t\rightarrow t+1}\right\|_1,\ 
\left\|\hat{\mathbf{q}}_{t\rightarrow t+1}+\mathbf{q}_{t\rightarrow t+1}\right\|_1\right).
\label{eq:pose_loss_pair}
\end{equation}
For depth, we use a masked regression loss over valid pixels, optionally weighted by $\hat{\mathbf{C}}$, together with a mild smoothness regularizer; we denote the resulting objective as $\mathcal{L}_{\text{depth}}$ and defer its exact form to implementation details.

The geometry losses augment the tokenizer objective in \cref{sec:tok_loss}:
\begin{equation}
\mathcal{L}
=
\mathcal{L}_{\text{tok}}
+
\lambda_{\text{geo}}
\left(
\lambda_{\text{depth}}\mathcal{L}_{\text{depth}}
+
\lambda_{\text{pose}}\mathcal{L}_{\text{pose}}
\right).
\label{eq:full_loss_simple}
\end{equation}

\subsubsection{Multi-Codebook Quantization}
\label{sec:mcb}

Joint supervision can create a capacity competition at the discrete bottleneck, where the same set of patch tokens must preserve appearance details while also carrying semantic representation and geometric cues.
Inspired by UniTok~\cite{tok2025.1} to increase capacity without changing the tokenization resolution, we extend the quantizer to a multi-codebook formulation.
Here, tokenization resolution refers to the number of patch tokens per frame and their spatial layout (the fixed $H_p\times W_p$ patch grid, equivalently $L$ tokens per frame).

Given a pre-quantization feature $\mathbf{z}_{t,l}\in\mathbb{R}^{d_q}$ for patch $l$ at time $t$, we first produce $M$ head-specific vectors using an attention-based splitter:
\begin{equation}
\mathbf{V}_{t,l}
=
P_{\text{attn}}(\mathbf{z}_{t,l})
\in
\mathbb{R}^{M d_q},
\qquad
\mathbf{v}^{(m)}_{t,l}\in\mathbb{R}^{d_q},
\label{eq:mcb_split}
\end{equation}
where $\mathbf{V}_{t,l}$ is reshaped into $\{\mathbf{v}^{(m)}_{t,l}\}_{m=1}^{M}$.
Each head is then quantized with an independent codebook $\mathcal{C}^{(m)}=\{\mathbf{c}^{(m)}_{k}\}_{k=1}^{K_m}$ using the same hard nearest-neighbor assignment as in \cref{sec:vq}, producing selected codewords $\mathbf{e}^{(m)}_{t,l}$.

We fuse the $M$ selected codewords into a single embedding passed to the shared post-encoder and decoders:
\begin{equation}
\mathbf{e}_{t,l}
=
P_{\text{merge}}\!\left([\mathbf{e}^{(1)}_{t,l};\dots;\mathbf{e}^{(M)}_{t,l}]\right)
\in \mathbb{R}^{d_q}.
\label{eq:mcb_merge}
\end{equation}
We found the attention-based split-and-merge along with the EMA updating helpful for balancing codebook utilization under joint objectives.

This design yields $M$ discrete indices per patch while keeping the patch grid unchanged, effectively increasing the representational capacity available to the tokenizer.

\subsection{Token-Based Planning Decoder}
\label{sec:planner}

To evaluate whether the learned discrete tokens can be consumed by downstream planning, we freeze the tokenizer from \cref{sec:tokenizer} and train a simple trajectory decoder on top of it.
At time $t$, the decoder takes the current tokens tokenized from RGB observation $\mathbf{I}_t$ and an ego status vector $\mathbf{s}_t\in\mathbb{R}^{11}$, and output a future ego trajectory
$\mathbf{Y}=\{\mathbf{y}_\tau\}_{\tau=1}^{T}$ with $\mathbf{y}_\tau=(x_\tau,y_\tau,\psi_\tau)\in\mathbb{R}^{3}$.
We use a fixed planning horizon and sampling interval so that $T$ is constant across all experiments.

We obtain patch-level token features $\mathbf{P}_t\in\mathbb{R}^{L\times d_p}$ from the frozen tokenizer and compress them into a small set of scene tokens using learnable registers and a small transformer:
\begin{equation}
\mathbf{Z}^{\text{scene}}_t = f_{\text{scene}}(\mathbf{P}_t)\in\mathbb{R}^{R\times d}.
\label{eq:scene_tokens}
\end{equation}
The ego status is embedded into $\mathbf{e}_t\in\mathbb{R}^{d}$.
A shallow attention module conditions $\mathbf{e}_t$ on $\mathbf{Z}^{\text{scene}}_t$, and an MLP head then predicts the multiple trajectories.
This decoder is intentionally lightweight, as our goal is to probe the quality of the token representation rather than to optimize planner architecture.
We train the trajectory head with a standard L1 regression objective, and only the trajectory closest to the ground-truth is supervised.

We add a simple scoring head that predicts PDM-style metric outcomes from the predicted trajectory and scene tokens~\cite{dataset2024.1}.
The supervision targets are obtained by running a rule-based evaluator on the predicted trajectory, and the score head is trained with binary cross-entropy.
This auxiliary loss provides a lightweight way to incorporate safety and compliance signals without changing the planner structure.
At inference time, we use the trajectory with the highest score as the plan.

We report planning performance under this fixed decoder to quantify token consumability; since the decoder capacity and training protocol are kept unchanged across tokenizers, differences primarily reflect the quality of the token representation.

\subsection{Autoregressive World Model over Discrete Tokens}
\label{sec:worldmodel}

We freeze the tokenizer and train a GPT-style autoregressive Transformer on the resulting discrete token sequences using next-token prediction~\cite{dl2020.1, tok2024.1}.

For each frame $t$, the tokenizer produces $L$ discrete tokens $\mathbf{k}_t\in\{1,\dots,K\}^{L}$.
We linearize the per-frame patch tokens with a fixed scan order and concatenate them over a temporal window, yielding a token stream
$\mathbf{x}_{1:S}$ with $S=T_{\text{win}}\cdot L$.
The world model predicts each token from its preceding context, and can be conditioned on driving signals $\mathbf{c}$ such as actions and ego states.

Conditioning is injected through adaptive layer normalization (AdaLN)~\cite{dl2023.1}.
In each Transformer block, we modulate the normalized hidden activations using scale and shift parameters predicted from a condition embedding.
The model is trained with teacher forcing and the standard cross-entropy loss over the discrete token vocabulary.
\section{Experiments}
\label{sec:exp}

\subsection{Implementation Details}
\label{sec:impl}

We summarize the key settings required to reproduce our experiments, and defer additional architectural and optimization details to Supplementary Material.

\textbf{Tokenizer variants.}
We train three tokenizers end-to-end from scratch under the same patching scheme: a naive reconstruction tokenizer (RGB reconstruction only), a DINO-guided tokenizer (adding frozen feature reconstruction with \texttt{DINOv3-B}~\cite{dl2025.1}), and a semantic representation-guided and geometry-enhanced tokenizer (further adding adjacent-frame depth and relative-pose supervision).
The first two use a single codebook of size $16384$, while the geometry-enhanced tokenizer uses $4$ codebooks of size $4096$ each.

\textbf{Downstream evaluation.}
For planning, we freeze the tokenizer and train the same lightweight (20M parameters) trajectory readout head on token features (\cref{sec:planner}), using the geometry-enhanced tokenizer by default in our main planning comparisons.
For world modeling, we train a GPT-style next-token Transformer (1B parameters) with cross-entropy on token sequences, using the semantic representation-guided tokenizer by default.

\subsection{Dataset}
\label{sec:dataset}

We conduct all experiments on the NAVSIM benchmark~\cite{dataset2024.1}, a data-driven non-reactive simulation and evaluation suite for autonomous driving.
NAVSIM is resampled from OpenScene~\cite{dataset2023.1}, which curates about 120 hours of driving logs from nuPlan~\cite{dataset2021.1}.
Following the benchmark construction, simple scenarios such as long straight driving are reduced, which strengthens the evaluation sensitivity for planning models~\cite{dataset2024.1,dataset2023.1}.

We follow a split protocol that separates token learning from downstream planning for fair comparison.
The tokenizer and the autoregressive world model are trained on the OpenScene training set.
The planning decoder is trained on the NAVSIM training set to match the standard planning benchmark setup.
All reported results are evaluated on the NAVSIM test split, and this test split does not overlap with the training data used above.
We use the Predictive Driver Model Score (PDMS) of NAVSIM to summarize planning quality~\cite{dataset2024.1}.
It aggregates five factors including No At-Fault Collision (NC), Drivable Area Compliance (DAC), Time-to-Collision (TTC), Comfort (Comf.), and Ego Progress (EP).

For cross-frame geometric supervision, we use absolute depth annotations derived from DVGT~\cite{dataset2025.1}, which aligns depth pseudo-labels with radar measurements and applies post-processing to filter noisy estimates, yielding absolute depth targets that we use as supervision for the tokenizer geometry branch.
These depth targets are only used during tokenizer training and are not required at inference time.

\subsection{Tokenizer Quality}
\label{sec:exp_tok_quality}
\begin{table}[tb]
  \caption{
  Baseline comparison on the NAVSIM test split.
  CB denotes the codebook configuration.
  $\Delta_{\mathrm{img}}^{\cos}$ and $\Delta_{\mathrm{img}}^{\mathrm{rms}}$ measure cosine distance and RMSE in the DINO feature space between reconstructed and ground-truth images. LlamaGen is the tokenizer used by DrivingGPT~\cite{baseline2025.2}.
  }
  \label{tab:tokenizer_comparison}
  \centering
  \setlength{\tabcolsep}{2.4pt}
  \renewcommand{\arraystretch}{1.03}
  \begin{tabular}{@{}l|c|ccc|cc@{}}
    \toprule
    Tokenizer & CB & rFID$\downarrow$ & PSNR$\uparrow$ & SSIM$\uparrow$ &
    $\Delta_{\mathrm{img}}^{\cos}\downarrow$ & $\Delta_{\mathrm{img}}^{\mathrm{rms}}\downarrow$ \\
    \midrule
    LlamaGen~\cite{baseline2025.2,tok2024.1}    & 16384          & 5.67 & 23.09 & 0.652 & 0.0869 & 0.167 \\
    Orbis~\cite{baseline2025.1}        & 2$\times$16384 & 5.53 & 25.94 & 0.773 & 0.0595 & 0.139 \\
    Ours-Rep+Geo & 4$\times$4096  & 5.14 & 26.33 & 0.769 & 0.0563 & 0.136 \\
    Ours-Rep     & 16384          & \textbf{4.15} & \textbf{26.51} & \textbf{0.774} & \textbf{0.0453} & \textbf{0.122} \\
    \bottomrule
  \end{tabular}
\end{table}

\begin{table}[tb]
  \caption{
  Tokenizer roadmap on the NAVSIM test split.
  AbsRel and $\delta_1$ summarize depth prediction; Trans and Rot report average relative-pose translation error in meters and rotation error in degrees.
  $\Delta_{\mathrm{dec}}^{\cos}$ and $\Delta_{\mathrm{dec}}^{\mathrm{rms}}$ measure cosine distance and RMSE between decoded DINO features and the frozen DINO targets.
  }
  \label{tab:ablation_sem_geo}
  \centering
  \setlength{\tabcolsep}{2.1pt}
  \renewcommand{\arraystretch}{1.03}
  \begin{tabular}{@{}l|cc|cc|cc|c@{}}
    \toprule
    Var. & AbsRel$\downarrow$ & $\delta_1\uparrow$ &
    Trans$\downarrow$ & Rot$\downarrow$ &
    $\Delta_{\mathrm{dec}}^{\cos}\downarrow$ & $\Delta_{\mathrm{dec}}^{\mathrm{rms}}\downarrow$ &
    PSNR$\uparrow$ \\
    \midrule
    Naive & --     & --     & --    & --   & --      & --      & 25.96 \\
    +Rep & --     & --     & --    & --   & \textbf{0.0346}  & \textbf{0.105}   & \textbf{26.51} \\
    +Geo & 0.0640 & 0.959  & 0.717 & 2.02 & 0.0717  & 0.153   & 23.90 \\
    +MCB & \textbf{0.0556} & \textbf{0.965}  & \textbf{0.647} & \textbf{2.01} & 0.0486  & 0.124   & 26.33 \\
    \bottomrule
  \end{tabular}
\end{table}
\subsubsection{Tokenizer Comparison} \label{sec:exp_tok_results}
We benchmark against representative discrete tokenizers that have been adopted in autonomous driving works. They used similar configurations for comparison, both using a resolution of $288\times512$~\cite{baseline2025.2,baseline2025.1}.
All results are reported on the NAVSIM test split; Orbis results are obtained by running the official checkpoint under the same evaluation pipeline.
\cref{tab:tokenizer_comparison} shows that Ours-Rep improves appearance reconstruction (rFID, PSNR, SSIM) over both baselines.
In addition, Ours-Rep achieves the lowest $\Delta_{\mathrm{img}}^{\mathrm{cos}}$ and $\Delta_{\mathrm{img}}^{\mathrm{rms}}$, indicating that reconstructed images remain closer to the ground truth in the frozen DINO feature space.
This diagnostic complements pixel-level metrics by evaluating reconstruction quality in the foundation feature space, which emphasizes object-level semantics and spatial layout (e.g., vehicles, road structure, and their relative arrangement) and is less sensitive to low-level appearance variations.
This aligns with our goal of representation-guided token learning in \cref{sec:tokenizer}, where tokens should retain semantically meaningful and structurally consistent cues for downstream modeling and planning.

Beyond these scalar reconstruction metrics, the compared tokenizers also differ in what can be decoded.
Our tokenizer includes a dedicated decoder to DINO patch features; the geometry-enhanced variant additionally supports depth and relative-pose decoding.

Taken together, the consistent gains in rFID, PSNR, and SSIM, along with the reduced DINO image-feature discrepancy, provide a coherent reconstruction diagnosis: the tokens support not only pixel-level fidelity but also closer structural consistency under a frozen representation used for supervision in \cref{sec:tokenizer}.

\subsubsection{Tokenizer Roadmap}
\label{sec:exp_tok_roadmap}
\cref{tab:ablation_sem_geo} summarizes the key trade-offs observed along our roadmap.
Adding representation supervision improves RGB reconstruction and yields substantially smaller decoded-feature discrepancies in the DINO space ($\Delta_{\mathrm{dec}}^{\mathrm{cos}}$/$\Delta_{\mathrm{dec}}^{\mathrm{rms}}$), indicating that the discrete tokens better preserve the target representation when it is directly supervised.
When geometric supervision is enabled, the tokens become predictive of depth and relative pose, reflected by improved depth metrics (AbsRel, $\delta_1$) and standard pose summaries (Trans in meters and Rot in degrees).
At the same time, PSNR drops noticeably, suggesting that a fixed discrete bottleneck can face a capacity trade-off when it is asked to preserve appearance, match a strong representation, and encode geometry cues simultaneously.
Finally, multi-codebook quantization mitigates this trade-off: it largely restores reconstruction quality while keeping depth and pose performance competitive, supporting the view that the earlier degradation is driven by limited discrete capacity rather than by conflicting objectives.

\begin{figure}[tb]
  \centering
  \includegraphics[width=\linewidth, trim=0mm 112mm 0mm 0mm, clip]{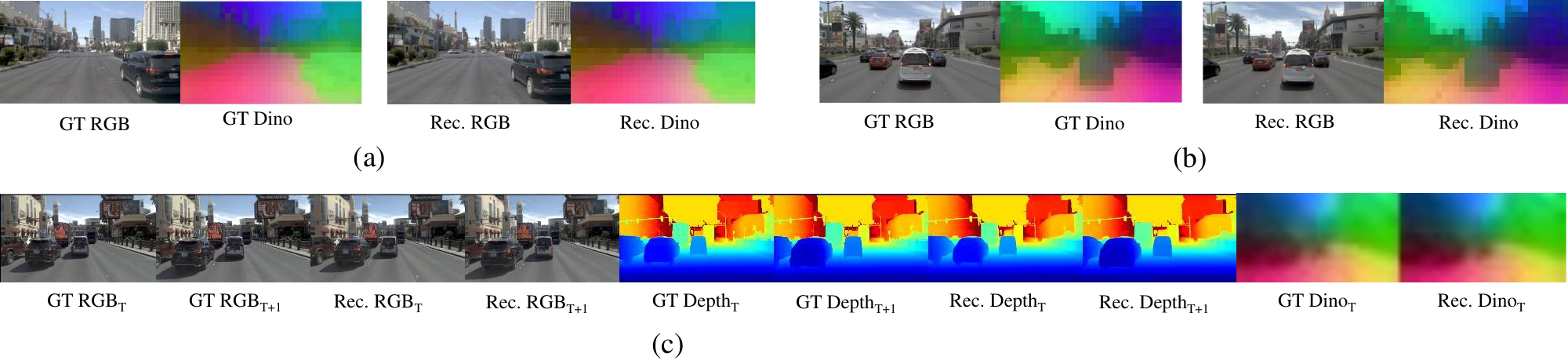}
  \caption{
    Qualitative visualization of tokenizer reconstructions.
    Subfigures (a) and (b) show the semantic representation-guided tokenizer, while (c) further incorporates geometry enhancement with a multi-codebook design.
    DINO features are visualized via PCA.
  }
  \label{fig:3}
\end{figure}

Overall, \cref{tab:ablation_sem_geo} indicates that the main challenge is not whether geometry supervision is compatible with representation alignment, but whether the discrete bottleneck has sufficient capacity to accommodate them simultaneously. 
\cref{fig:3} shows the visualization of the reconstruction results.
This observation motivates using the geometry-enhanced tokenizer as the default choice when probing planning readout in the next experiment, where we keep the downstream planner lightweight and focus on how much geometric and structural information is recoverable from the tokens.

\subsection{Planning Performance of Driving Tokens}
\begin{table}[tb]
  \caption{
    Comparison on the NAVSIM test split. Methods marked with $^{\dagger}$ use visual tokens from a frozen tokenizer as input; the remaining methods are end-to-end.
    Sensor setting: C denotes multi-camera, C+L denotes cameras with LiDAR, and *C denotes the single-view (1V) setting.
  }
  \label{tab:navsim_roadmap}
  \centering
  \setlength{\tabcolsep}{3.2pt}
  \renewcommand{\arraystretch}{1.05}
  \begin{tabular}{@{}l|c|ccccc|c@{}}
    \toprule
    Method & Sensor & NC$\uparrow$ & DAC$\uparrow$ & TTC$\uparrow$ & Comf.$\uparrow$ & EP$\uparrow$ & PDMS$\uparrow$ \\
    \midrule
    VADv2~\cite{e2e2024.2}              & C   & 97.2 & 89.1 & 91.6 & 100.0 & 76.0 & 80.9 \\
    UniAD~\cite{e2e2023.1}              & C   & 97.8 & 91.9 & 92.9 & 100.0 & 78.8 & 83.4 \\
    Para-Drive~\cite{e2e2024.1}         & C   & 97.9 & 92.4 & 93.0 & 99.8  & 79.3 & 84.0 \\
    TransFuser~\cite{e2e2022.1}         & C+L & 97.7 & 92.8 & 92.8 & 100.0 & 79.2 & 84.0 \\
    DRAMA~\cite{e2e2024.4}              & C+L & 98.0 & 93.1 & 94.8 & 100.0 & 80.1 & 85.5 \\
    DiffusionDrive~\cite{e2e2025.1}     & C+L & 98.2 & 96.2 & 94.7 & 100.0 & 82.2 & 88.1 \\
    WoTE~\cite{e2e2025.3}               & C+L & 98.5 & 96.8 & 94.9 & 99.9  & 81.9 & 88.3 \\
    ResWorld~\cite{e2e2026.1}           & C   & 98.9 & 96.5 & 95.6 & 100.0 & 83.1 & 89.0 \\
    Hydra-MDP++~\cite{e2e2025.7}        & C+L & 98.6 & 98.6 & 95.1 & 100.0 & 85.7 & 91.0 \\
    DriveSuprim~\cite{e2e2025.8}        & C   & 98.6 & 98.6 & 95.5 & 100.0 & 91.3 & \textbf{93.5} \\
    \midrule
    Epona$^{\dagger}$~\cite{e2e2025.2}          & *C & 97.9 & 95.1 & 93.8 & 99.9  & 80.4 & 86.2 \\
    PWM$^{\dagger}$~\cite{e2e2025.4}            & *C & 98.6 & 95.9 & 95.4 & 100.0 & 81.8 & 88.1 \\
    DriveLaW$^{\dagger}$~\cite{e2e2025.6}       & *C & 99.0 & 97.1 & 96.7 & 100.0 & 81.3 & 89.1 \\
    DriveVLA-W0$^{\dagger}$~\cite{e2e2025.5}    & *C & 98.7 & 99.1 & 95.3 & 99.3  & 83.3 & 90.2 \\
    Ours$^{\dagger}$                            & *C & 98.7&	98.2&	95.9 &	100.0	& 87.3 & \textbf{91.8} \\
    \bottomrule
  \end{tabular}
\end{table}

\begin{table}[tb]
  \caption{
    Tokenizer Ablation on the NAVSIM test split.
  }
  \label{tab:ablation_rep_geo_mcb}
  \centering
  \setlength{\tabcolsep}{3.2pt}
  \renewcommand{\arraystretch}{1.05}
  \begin{tabular}{@{}c|ccc|ccccc|c@{}}
    \toprule
    ID & Rep & Geo & MCB &
    NC$\uparrow$ & DAC$\uparrow$ & TTC$\uparrow$ & Comf.$\uparrow$ & EP$\uparrow$ &
    PDMS$\uparrow$ \\
    \midrule
    1 & $\times$       & $\times$       & $\times$       & 98.2 & 94.8 & 94.6 & 100.0 & 77.8 & 85.5 \\
    2 & $\checkmark$   & $\times$       & $\times$       & 98.7 & 97.4 & 96.3 & 100.0 & 82.1 & 89.4 \\
    3 & $\checkmark$   & $\checkmark$   & $\times$       & 98.6 & 97.6 & 95.7 & 100.0 & 86.3 & 90.9 \\
    {4} & {$\checkmark$} & {$\checkmark$} & {$\checkmark$} &
    98.7&	98.2&	95.9 &	100.0	& 87.3 & \textbf{91.8} \\
    \bottomrule
  \end{tabular}
\end{table}

\begin{figure}[tb]
  \centering
  \includegraphics[width=\linewidth, trim=10mm 130mm 8mm 0mm, clip]{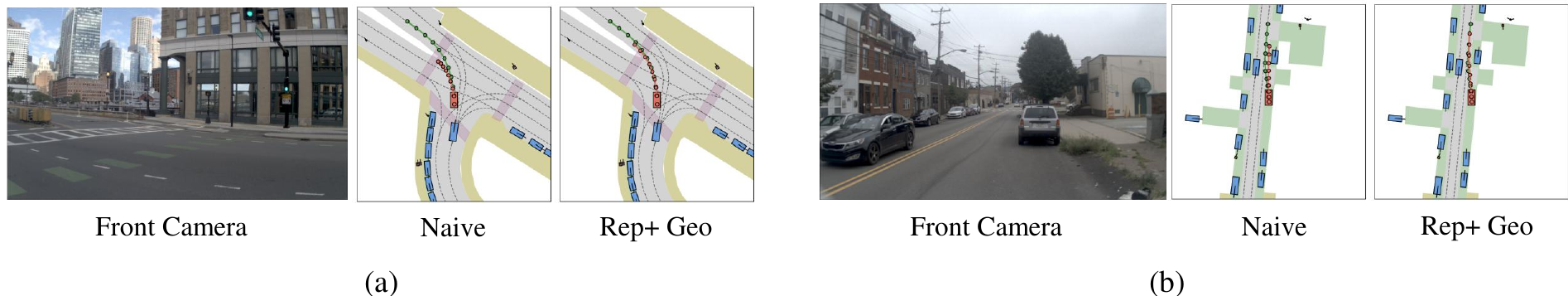}
  \caption{
    Qualitative visualization of tokenizer ablation for planning.
    Subfigures (a) and (b) show two cases that the naive tokenizer fails, while the Rep+Geo tokenizer succeeds.
  }
  \label{fig:4}
\end{figure}

We evaluate driving-token quality through planning on the NAVSIM test split.
We report PDMS and its standard components (NC, DAC, TTC, Comfort, and EP) under the official evaluation pipeline.
\subsubsection{Comparisons with State-of-the-Art Methods}
To place our results in context, we compare against two groups of methods in \cref{tab:navsim_roadmap}:
end-to-end planning systems that learn perception and planning jointly, and token-based methods that take visual tokens from a frozen tokenizer as input.
\cref{tab:navsim_roadmap} reports planning performance on the NAVSIM test split.
Under the single-view camera setting, our token-based planner achieves the highest PDMS among methods that plan from frozen visual tokens.
Across sensor settings, several end-to-end systems benefit from multi-camera or LiDAR inputs, which makes direct comparison less controlled, but the results indicate that the learned tokens support competitive single-view planning when used as frozen inputs.


\subsubsection{Tokenizer Ablation for Planning}

The ablation in \cref{tab:ablation_rep_geo_mcb} follows the tokenizer roadmap and isolates how each design choice affects planning readout under the same decoder and training protocol.
Adding representation supervision improves PDMS substantially over the naive reconstruction-only tokenizer, with consistent gains on NC, DAC, TTC, and EP.
Enabling geometric supervision further increases PDMS, mainly through EP, which is consistent with the intent of making tokens more predictive of state and motion cues used by planning.
Finally, multi-codebook quantization yields the best overall PDMS while preserving strong safety and compliance components, suggesting that additional discrete capacity helps retain both representation and geometry information that the readout can use. Furthermore, the visualized case comparisons in \cref{fig:4} also reflect the above conclusions.



\subsection{Visual Generation of the World Model}
\label{sec:wm_gen}
\begin{figure}[tb]
  \centering
  \includegraphics[width=\linewidth, trim=0mm 48mm 0mm 0mm, clip]{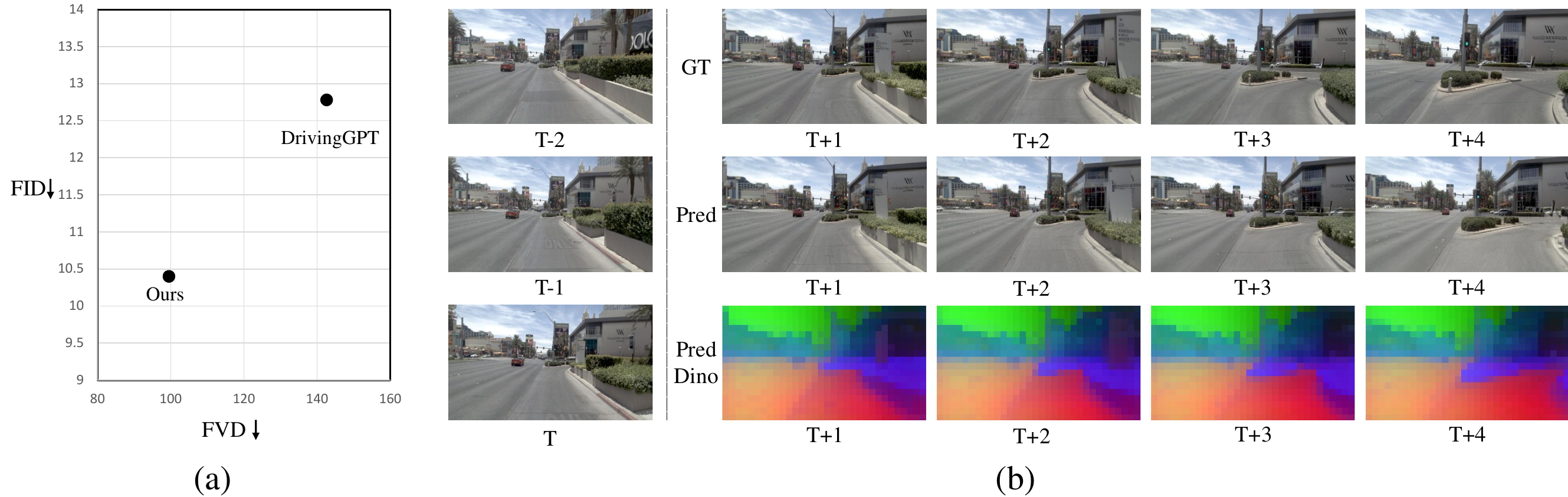}
  \caption{
    Generative performance of world models.
    Subfigures (a) compares the generation performance of our method with the baseline. Subfigures (b) visualizes the generation results.
  }
  \label{fig:5}
\end{figure}

Following DrivingGPT~\cite{baseline2025.2}, we evaluate visual rollout using a fixed history-and-prediction protocol.
We condition the world model on the past three frames and generate the next eight frames at 2\,Hz.
All metrics are computed on the NAVSIM test split at the same image resolution of $288 \times 512$.

As shown in \cref{fig:5}, the world model trained on our tokens achieves better FID and FVD than the baseline under the matched setting.
This indicates that our tokenizer does not sacrifice generative usability when optimized for representation guidance and downstream planning probes.

\cref{fig:5} visualizes a representative rollout example.
The model generates coherent future frames and, through the same discrete tokens, we can also decode the corresponding DINO feature targets for each predicted frame.
The decoded DINO features remain consistent with the generated images, providing an additional view of structural consistency along the rollout.

Overall, these results support the premise in Sec.~\ref{sec:intro} that a single discrete tokenization can serve both planning readout and world model generation within a unified training and evaluation pipeline.

\section{Conclusion}
\label{sec:conclusion}
We studied discrete visual tokenization for autonomous driving under the requirement that the same tokens should support planning readout and autoregressive sequence prediction. We proposed a tokenizer trained with representation supervision from frozen DINO features, complemented by RGB reconstruction with perceptual and adversarial objectives. We further introduced adjacent-frame depth and relative-pose supervision to encourage geometry-aware tokens, and used multi-codebook quantization to reduce capacity pressure under joint training. Across NAVSIM evaluations, the resulting tokens improve reconstruction fidelity and yield smaller discrepancies in the DINO feature space, which serves as a practical diagnostic of structural consistency beyond pixel metrics. Using fixed downstream models, the tokens enable strong planning performance and support high-quality world model rollouts. Future work includes extending geometric supervision beyond adjacent pairs and unifying planning and world modeling within a unified architecture.



%
%
\bibliographystyle{splncs04}
\bibliography{main}
\end{document}